\pdfoutput=1

\documentclass[11pt]{article}

\usepackage{EMNLP2022}

\usepackage{times}
\usepackage{latexsym}
\usepackage{graphicx}
\usepackage{booktabs}
\usepackage{dirtytalk}
\usepackage[inline]{enumitem}

\usepackage[T1]{fontenc}

\usepackage[utf8]{inputenc}

\usepackage[shrink=50]{microtype}

\usepackage{inconsolata}
\usepackage{cleveref}

\title{A Major Obstacle for NLP Research: Let's Talk about Time Allocation!}

\author{
    Katharina Kann${ }^{\spadesuit}$ \and Shiran Dudy${ }^{\spadesuit}$  \and Arya D. McCarthy${ }^{\clubsuit}$ \\
    ${ }^{\spadesuit}$University of Colorado Boulder\\\texttt{firstname.lastname@colorado.edu}
    \\${ }^{\clubsuit}$Johns Hopkins University\\\texttt{lastname@jhu.edu}
  }

\begin{document}
\maketitle
\begin{abstract}
The field of natural language processing (NLP) has grown over the last few years: conferences have become larger, we have published an incredible amount of papers, and state-of-the-art research has been implemented in a large variety of customer-facing products. 
However, this paper 
argues that we have been less successful than we \textit{should} have been 
and reflects on where and how the field fails to tap its full potential. 
Specifically, we demonstrate that, in recent years,  \textbf{subpar time allocation has been a major obstacle for NLP research}. We outline multiple concrete problems together with their negative consequences and, importantly, suggest remedies to improve the status quo. We hope that this paper will be a starting point for discussions around which common practices are -- or are \textit{not} -- beneficial for NLP research.
\end{abstract}

\section{Introduction}
\textit{Why did I get nothing done today?} 
is a question many people ask themselves frequently throughout their professional careers.
Psychologists agree on good time management skills being of utmost importance for a healthy and productive lifestyle \cite[\textit{inter alia}]{lakei1973get,claessens2007review,major2002work}. However, many academics and industry researchers lack time management skills, working long days and getting not enough done -- not even the interesting experiment they had wanted to start over a year ago.

In this position paper, we argue that natural language processing (NLP) as a field has a similar problem: we do not allocate our time well. Instead, we spend it on things that seem more urgent than they are, are easy but unimportant, or result in the largest short-term gains. This paper identifies the largest traps the authors believe the NLP community falls into. We then provide, for each of the four identified problems (P1--P4), suggested remedies. While we know that -- just as for individuals -- change takes time, we hope that this paper, in combination with the EMNLP 2022 special theme \textit{Open questions, major obstacles, and unresolved issues in NLP}, will ignite
critical 
discussions.

\begin{figure}[t]
\centering
\includegraphics[width=.98\columnwidth]{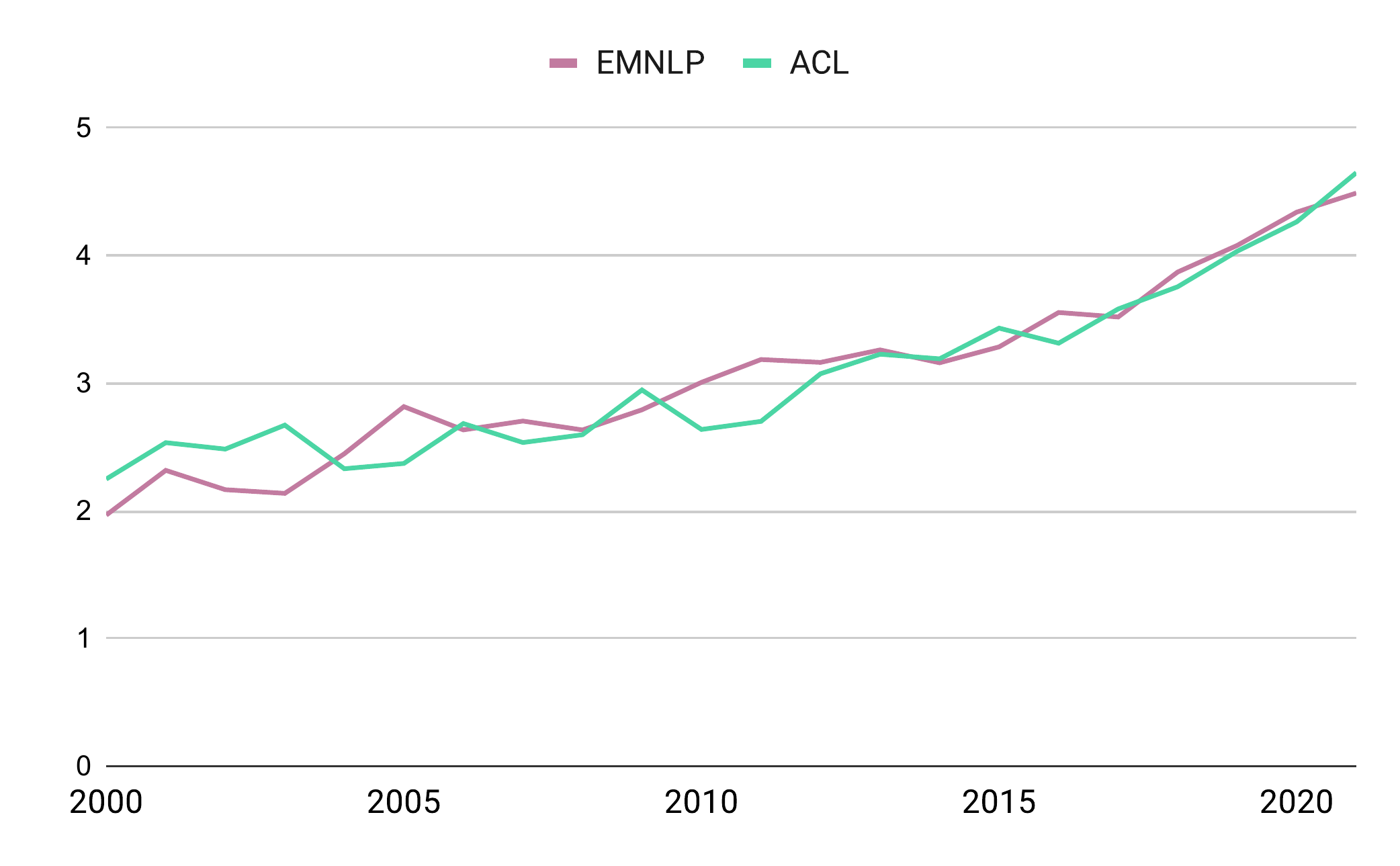}
\caption{Avg. \# of authors per paper; 2000--2021.\label{fig:authors}}
\end{figure}
\paragraph{Related Work}
Over the last couple of years, multiple papers have provided critical reflections on the state of affairs in NLP research: \citet{bender-koller-2020-climbing} criticizes the hype around language models and argues, similarly to \citet{bisk-etal-2020-experience}, that true understanding is impossible when language is detached from the physical world. In contrast, \citet{bowman-2022-dangers} talks about the risks associated with \textit{underclaiming}.
Turning to evaluation, \citet{bowman-dahl-2021-will} provides a critical view on benchmarking, and \citet{rodriguez-etal-2021-evaluation} proposes ways to improve leaderboards in order to truly track progress. Other position papers discuss the importance of data curation \cite{rogers-2021-changing} and the need for focusing on the user for natural language generation \cite{dudy-etal-2021-refocusing,flek2020returning}. \citet{bianchi-hovy-2021-gap} identifies general concerning trends in NLP research.
\citet{parcalabescu-etal-2021-multimodality} discusses our use of the term \textit{multimodality} and proposes to use \textit{task-specific} definitions of multimodality 
in the machine learning era.
\citet{church_2020}
 discusses downward trends in reviewing quality and whether these can be mitigated.
 We add to those meta-level papers by discussing 
 subpar use of time
 as a major problem.
 
\section{What Is Going Wrong?}
\subsection{P1: Too Many Papers per Author}

\paragraph{The Situation} 
Publications in NLP are cheap compared to many other fields: there is no need to set up complicated real-world experiments (as, e.g., in physics),
existing data can be used for many studies, and lately even much of the code we use is readily available. Thus, the time from idea to final paper can be extremely short.
Some researchers also split one substantial paper's work into 2--5 less dense and structurally similar papers.

Consequently, NLP researchers publish a lot: Rei\footnote{\url{https://www.marekrei.com/blog/ml-and-nlp-publications-in-2021/}} finds that the 14 most productive first authors in NLP published 9 (1 researcher), 6 (2 researchers), and 5 (11 researchers) papers in 2021. And this number only counts the \textit{most prestigious conferences in NLP}: Google Scholar shows that, across all venues, the first 3 authors published 16, 7, and 7 papers. 

While some enjoy writing, many -- especially junior -- NLP researchers feel external pressure to publish in large volumes; quantity often overshadows quality of publications for hiring decisions, and PhD applicants struggle to find advisors if they do not have multiple relevant publications. 

\paragraph{Negative Consequences} 
A straightforward consequence of the pressure to publish is that \textit{much of an NLP researcher's time goes into writing}: 
conservatively assuming one week of full-time writing per paper, the authors with the most papers respectively spend 16, 7, and 7 weeks per year just writing; this is nearly $\frac{1}{3}$ of the most productive author's year.

The second negative consequence is the \textit{time needed to review this many papers}: reviewing one substantial paper would be quicker than reviewing 5 separate ones, especially if reviewers are not shared. This lowers review quality, frustrates authors, and causes errors to be missed. The latter then misinforms other researchers, also wasting their time.

Third, the ongoing race for publications makes it \textit{difficult for researchers to stop and reflect on if what they are currently working on is worthwhile}. It also leads to \textit{mixed feelings regarding the start of ambitious, high-risk/high-reward research}: many researchers are scared away by the prospect of potentially not obtaining their expected outcomes and being unable to publish. 
Thus, the need to constantly produce large quantities of output not only reduces the quality of individual papers, but also hinders meaningful progress of the field by encouraging the pursuit of superficial research questions. 

Finally, \textit{thorough scholarship is extremely difficult} in this environment. This leads to all sorts of shortcomings in NLP publications --  missing references, mathematical errors, and even nonsensical experimental designs -- which are then overlooked by overworked reviewers \citep{church_2020}.

\paragraph{Suggested Remedies}
To change the state of the field, we can either change our expectations or the available opportunities. 
For the former, it is crucial that quality is valued more than quantity for hiring. To start, we recommend having reviews be publicly available (as done, e.g., by the Conference on Neural Information Processing Systems\footnote{\url{https://neurips.cc}}), to help people from adjacent fields understand the value of a candidate's publication. Another option is to standardize requesting reviews from experts (in addition to letters of recommendation). To reduce the opportunities for submitting large amounts of less impactful papers, we could set an upper limit for the number of (first-author) papers one can submit. This could be a hard limit or a soft limit with a penalty for too many low-quality submissions, such as blocking papers with low average scores from resubmission for a fixed period of time.\footnote{This is current practice in TACL but not at conferences.} 

\subsection{P2: Too Many Authors per Paper}
\paragraph{The Situation}
The second problem we highlight is the inverse of the first: too many authors per paper,\footnote{Examples with >20 authors are \citet{nan-etal-2021-dart}, \citet{srivastava2022beyond}, and \citet{gehrmann-etal-2021-gem,gemv2}.}
given the strategies we employ to manage collaborations.
As shown in Figure \ref{fig:authors}, author lists are, on average, becoming longer and longer: in 2000, the average number of authors on ACL and EMNLP papers was 2.25 and, respectively, 1.97, but that number had increased to 4.65 and, respectively, 4.49 in 2021. 
Large collaborations can greatly advance science and, if done well, are beneficial to all participating researchers.
However, they also pose an unintended challenge: many times, each author's expected, as well as actual, contribution becomes unclear. The former is often a consequence of a lack of communication or team management skills. The latter is the result of 
NLP not having a standardized way to communicate each researcher's contribution to collaborative projects.

In a traditional two-author setting with a student and their advisor, it is generally understood that the student does most of the hands-on work and their advisor guides the research and writing process.
However, with more authors, the situation becomes less clear to both authors and readers.

\paragraph{Negative Consequences}
When expected contributions are unclear to the authors themselves, it is easy to have \textit{too many cooks spoil the broth}: e.g., one author could write one section while 
one of their colleagues rewrites another section in a way that makes combining them non-trivial and time-consuming. 
Additionally, being vague about each author's contributions can lead to \textit{
friction around authorship}, which take time as well as mental energy and a toll on the relationship between two people; also, authorship discussions tend to disadvantage members of underrepresented groups \cite{doi:10.1126/sciadv.abe4639}. 

Worse, however, is a situation in which it is the reader to whom it is not obvious what each authors' contribution has been. Some researchers giving authorship to people whose contribution was minimal \textit{devalues the time and work of middle authors who actually do contribute a lot}.

Another problem with too many authors is that \textit{miscommunication easily wastes time and resources}. For instance, it is easy to be inconsistent if experiments are run by multiple researchers, who might not  use the same codebase.

\paragraph{Suggested Remedies} 
In order to avoid situations where the contributions of individual authors are unclear to the reader -- and, thus, accurate assignment of credit is impossible -- 
we propose a straightforward solution that can completely eliminate this negative consequence of large collaborations: publishing a contribution statement \cite{brand2015beyond} for each paper. This is common in other fields but very rare in NLP (a notable exception is, e.g., \citet{srivastava2022beyond}). Making a contribution statement mandatory for NLP publications would be easy but extremely effective.

For group management, setting expectations together and communicating the expected roles of all involved parties, including the possible authorship order can save time and energy toll.\footnote{\citet{moster2021my} offers insights on managing collaborations adjusted to remote work conditions.} We suggest that doing this right at the beginning of each collaborative project should become common practice in NLP ("\#EMNLP2022Rule"). However, it has been shown that many principal investigators (PIs) lack training in lab and personnel management skills \cite{van2018leadership}. Thus, PIs and their research groups would likely benefit from explicit training. One possible way to achieve this could be to extend existing mentoring initiatives at NLP conferences to focus more on leadership skills. Another suggestion mentioned by \citet{van2018leadership} -- which we recommend for NLP -- is that PIs should ask for feedback from their groups more regularly.

\subsection{P3: Gatekeeping}
\paragraph{The Situation} 
We do like unconventional topics (e.g., the connection between synesthesia and character-level NLP models \cite{kann-monsalve-mercado-2021-coloring}), 
and statements like "This work is too interdisciplinary to get accepted" or "This work would be better for a workshop on a specific topic" are hardly ever true. However, reviewers in NLP like papers that resemble those they themselves have previously published. They only accept non-mainstream submissions if they are written in a very specific style: authors need to know how to pitch a topic to the NLP community.

For readers new to publishing in NLP, here are the basic guidelines we have found for getting a paper accepted -- many of which are nonsensical: 
\begin{enumerate*}[label=\arabic*)]
    \item Your submitted paper should always have the exact maximum number of pages -- not a line more or less. 
    \item The first section should be called \textit{Introduction}. 
    \item The last section should be called \textit{Conclusion} -- not \textit{Discussion} or similar. 
    \item You should have a figure that is (somewhat) related to your paper's content on the top right corner of the first page.
    \item You should have equations in your paper -- complicated equations will increase your chances of acceptance \citep{lipton-steinhart}.
    \item Do not explicitly write out popular model equations, e.g., for the LSTM \cite{6795963}.
    \item The \textit{Related Work} section should come immediately before the \textit{Conclusion}, to make your novelty seem larger.
    \item Do not present only a dataset---provide empirical results, even if they are unimportant.
\end{enumerate*}

\paragraph{Negative Consequences} 
This
gatekeeping especially affects people whose research mentors 
are not able to teach them the style of the NLP community: 1) people from universities with little experience in NLP research, 2) researchers from countries not traditionally part of the international NLP community, and 3) people from adjacent fields, such as psychology, social science, or even linguistics. 

Thus, \textit{gatekeeping reinforces existing social inequalities and harms our research progress}, as we get exposed to  groundbreaking ideas later than necessary -- or never. It is also a huge waste of our time: for instance, there is no reason why content presented in 7.56 pages should be less impactful than content presented in 8 pages. However,
we, as a community, make it an issue and \textit{cause researchers to waste hours trimming or extending papers}. Similarly, we force people to \textit{waste their time thinking about which equations they can put into a paper that does not, in fact, benefit from them}. 

\paragraph{Suggested Remedies} 
We argue that resolving the problem of gatekeeping is crucial in order to allow our field to grow in a healthy way. We make two suggestions: 1) We need to explicitly educate reviewers to not take
superficial properties of papers into account.
This could be implemented, e.g., in the form of mandatory training videos for all ACL reviewers. 
However, this is a type of implicit bias \cite{greenwald1995implicit} and we encourage more discussion on possible solutions.  2) While we are waiting for this to be effective, we need to level the playing field by making unofficial rules and tricks widely known. The easiest way would be to publish explanations for first-time submitters together with calls for papers. Mentoring programs are great alternatives: while they are timewise costly for individuals, they will, in the long run, save time for the field as a whole.

\subsection{P4: Missing the Point}
\paragraph{The Situation}
NLP 
aims to build technology that improves the lives of its end users. However, 
NLP research is often purely technically driven, and actual human needs are investigated little or not at all \cite{flek2020returning,dudy-etal-2021-refocusing};
this is especially prevalent when building tools for communities speaking low-resource languages \cite{caselli-etal-2021-guiding}.
This can -- and does -- result in researchers focusing on irrelevant problems. 
A similar problem is what we call \textit{legacy research questions}: research questions that are motivated by problems or tools that are no longer relevant. Examples pointed out by \citet{bowman-2022-dangers} are papers motivated by the brittleness of question answering (QA) systems whose performance has long been surpassed by the state of the art or an analysis and drawing of conclusions based on outdated systems like BERT \cite{devlin-etal-2019-bert}.\footnote{It is, of course, possible to perform interesting studies involving older models. However, this requires well motivated research questions.} 

To quantify this problem, we performed a case study by randomly sampling and examining 30 papers from human-oriented tracks at EMNLP 2021.\footnote{The tracks we consider are: \textit{Machine translation and Multilinguality}, \textit{Dialogue and Interactive Systems}, \textit{Question Answering}, and \textit{Summarization}.} Only 3 papers engaged with users through evaluation and only 2 papers grounded their research questions in user needs; details can be found in Appendix \ref{sec:appendix}. 

Last, looking at recent top-tier conferences in the field of NLP, a substantial amount of papers focus on what we call \textit{quick research questions}, i.e., projects which maximize short-term gains for the researcher(s):
\citet{baden} identify that the majority of NLP research for text analysis is devoted to \say{easy problems}, instead of aiming to \say{measure much more demanding constructs.}

\paragraph{Negative Consequences} 
Work that is missing the point does not move the field in a meaningful direction. It \textit{wastes the researcher's time} by detracting from topics that truly benefit the community, the public, or the researcher themselves. 
Next, they \textit{waste the reviewers' time} as well as \textit{the general reader's time} by failing to provide insights. 
They also needlessly use computing resources, thus contributing to the climate crisis~\citep{strubell2019energy}.
Ignoring user needs further dangerously bears the risk of causing real harm to stakeholders \citep{raji2022fallacy}. Designing technology without the participation of potential users has in the past led to spectacular product failures \citep{Khari2021the,Simon2020google}. 

Finally, work on superficial research questions can be fast and result in a large amount of research output. In our current system that values quantity over quality for hiring, researchers working on superficial questions tend to have more successful careers. This, in turn, \textit{encourages new researchers to also waste their time} by doing something similar.

\paragraph{Suggested Remedies} 
It is important for NLP researchers to engage more with the intended users of the technology we build. This could be encouraged during the review process, e.g., with targeted questions. Legacy research questions will need to be detected during reviewing as well -- raising awareness of this phenomenon will likely reduce impacted submissions and acceptance of papers focused on legacy research questions alike.
Regarding quick research questions, one of the remedies suggested for P1 
could be a possible solution here as well: moving towards valuing quality over quantity.

\section{Conclusion}
In this paper, we outlined how several problematic practices in NLP research lead to a waste of the most important resource we have -- our time -- and, thus, constitute major obstacles for NLP research. We suggested multiple possible solutions to existing problems. We hope to foster much-needed discussion around how we, as a community, envision moving forward in the face of these concerns.

\section*{Limitations}
As we focus on time allocation, this is not an exhaustive list of problems we see in our research community. However, other concerns are 
beyond the scope of this work. Similarly, not all mentioned problems apply to all groups -- it is, for instance, totally possible that individual groups excel at managing large collaborations. 

We further do not claim that our suggested remedies are perfect solutions. They come with their own sets of challenges and should be implemented with care: for instance, contribution statements could unintentionally minimize contributions that do not make it into the final paper.
Additionally, we do not claim to have listed all possible remedies for the identified problems. By contrast, we explicitly encourage other researchers to start discussing ways to improve the status quo.

\section*{Acknowledgments}
We would like to thank the anonymous reviewers
for their thought-provoking comments as well as the members of University of Colorado Boulder's NALA Group for their helpful feedback. This research
was supported by the NSF National AI Institute
for Student-AI Teaming (iSAT) under grant DRL
2019805. The opinions expressed are those
of the authors and do not represent views of the
NSF.
ADM is supported by an Amazon Fellowship and a Frederick Jelinek Fellowship.

\bibliography{anthology,custom}

\begin{thebibliography}{63}
\expandafter\ifx\csname natexlab\endcsname\relax\def\natexlab#1{#1}\fi

\bibitem[{Amplayo et~al.(2021)Amplayo, Angelidis, and
  Lapata}]{amplayo2021aspect}
Reinald~Kim Amplayo, Stefanos Angelidis, and Mirella Lapata. 2021.
\newblock Aspect-controllable opinion summarization.
\newblock In \emph{Proceedings of the 2021 Conference on Empirical Methods in
  Natural Language Processing}, pages 6578--6593.

\bibitem[{Baden et~al.(2022)Baden, Pipal, Schoonvelde, and van~der
  Velden}]{baden}
Christian Baden, Christian Pipal, Martijn Schoonvelde, and Mariken A. C.~G
  van~der Velden. 2022.
\newblock \href {https://doi.org/10.1080/19312458.2021.2015574} {Three gaps in
  computational text analysis methods for social sciences: A research agenda}.
\newblock \emph{Communication Methods and Measures}, 16(1):1--18.

\bibitem[{Bara et~al.(2021)Bara, Sky, and Chai}]{bara2021mindcraft}
Cristian-Paul Bara, CH-Wang Sky, and Joyce Chai. 2021.
\newblock Mindcraft: Theory of mind modeling for situated dialogue in
  collaborative tasks.
\newblock In \emph{Proceedings of the 2021 Conference on Empirical Methods in
  Natural Language Processing}, pages 1112--1125.

\bibitem[{Bender and Koller(2020)}]{bender-koller-2020-climbing}
Emily~M. Bender and Alexander Koller. 2020.
\newblock \href {https://doi.org/10.18653/v1/2020.acl-main.463} {Climbing
  towards {NLU}: {On} meaning, form, and understanding in the age of data}.
\newblock In \emph{Proceedings of the 58th Annual Meeting of the Association
  for Computational Linguistics}, pages 5185--5198, Online. Association for
  Computational Linguistics.

\bibitem[{Bianchi and Hovy(2021)}]{bianchi-hovy-2021-gap}
Federico Bianchi and Dirk Hovy. 2021.
\newblock \href {https://doi.org/10.18653/v1/2021.findings-acl.340} {On the gap
  between adoption and understanding in {NLP}}.
\newblock In \emph{Findings of the Association for Computational Linguistics:
  ACL-IJCNLP 2021}, pages 3895--3901, Online. Association for Computational
  Linguistics.

\bibitem[{Bisk et~al.(2020)Bisk, Holtzman, Thomason, Andreas, Bengio, Chai,
  Lapata, Lazaridou, May, Nisnevich, Pinto, and
  Turian}]{bisk-etal-2020-experience}
Yonatan Bisk, Ari Holtzman, Jesse Thomason, Jacob Andreas, Yoshua Bengio, Joyce
  Chai, Mirella Lapata, Angeliki Lazaridou, Jonathan May, Aleksandr Nisnevich,
  Nicolas Pinto, and Joseph Turian. 2020.
\newblock \href {https://doi.org/10.18653/v1/2020.emnlp-main.703} {Experience
  grounds language}.
\newblock In \emph{Proceedings of the 2020 Conference on Empirical Methods in
  Natural Language Processing (EMNLP)}, pages 8718--8735, Online. Association
  for Computational Linguistics.

\bibitem[{Bowman(2022)}]{bowman-2022-dangers}
Samuel Bowman. 2022.
\newblock \href {https://doi.org/10.18653/v1/2022.acl-long.516} {The dangers of
  underclaiming: Reasons for caution when reporting how {NLP} systems fail}.
\newblock In \emph{Proceedings of the 60th Annual Meeting of the Association
  for Computational Linguistics (Volume 1: Long Papers)}, pages 7484--7499,
  Dublin, Ireland. Association for Computational Linguistics.

\bibitem[{Bowman and Dahl(2021)}]{bowman-dahl-2021-will}
Samuel~R. Bowman and George Dahl. 2021.
\newblock \href {https://doi.org/10.18653/v1/2021.naacl-main.385} {What will it
  take to fix benchmarking in natural language understanding?}
\newblock In \emph{Proceedings of the 2021 Conference of the North American
  Chapter of the Association for Computational Linguistics: Human Language
  Technologies}, pages 4843--4855, Online. Association for Computational
  Linguistics.

\bibitem[{Brand et~al.(2015)Brand, Allen, Altman, Hlava, and
  Scott}]{brand2015beyond}
Amy Brand, Liz Allen, Micah Altman, Marjorie Hlava, and Jo~Scott. 2015.
\newblock Beyond authorship: attribution, contribution, collaboration, and
  credit.
\newblock \emph{Learned Publishing}, 28(2):151--155.

\bibitem[{Caselli et~al.(2021)Caselli, Cibin, Conforti, Encinas, and
  Teli}]{caselli-etal-2021-guiding}
Tommaso Caselli, Roberto Cibin, Costanza Conforti, Enrique Encinas, and
  Maurizio Teli. 2021.
\newblock \href {https://doi.org/10.18653/v1/2021.nlp4posimpact-1.4} {Guiding
  principles for participatory design-inspired natural language processing}.
\newblock In \emph{Proceedings of the 1st Workshop on NLP for Positive Impact},
  pages 27--35, Online. Association for Computational Linguistics.

\bibitem[{Chen et~al.(2021)Chen, Ma, Chen, Dong, Zhang, Pan, Wang, and
  Wei}]{chen2021zero}
Guanhua Chen, Shuming Ma, Yun Chen, Li~Dong, Dongdong Zhang, Jia Pan, Wenping
  Wang, and Furu Wei. 2021.
\newblock Zero-shot cross-lingual transfer of neural machine translation with
  multilingual pretrained encoders.
\newblock In \emph{Proceedings of the 2021 Conference on Empirical Methods in
  Natural Language Processing}, pages 15--26.

\bibitem[{Church(2020)}]{church_2020}
Kenneth~Ward Church. 2020.
\newblock \href {https://doi.org/10.1017/S1351324920000030} {Emerging trends:
  Reviewing the reviewers (again)}.
\newblock \emph{Natural Language Engineering}, 26(2):245–257.

\bibitem[{Claessens et~al.(2007)Claessens, Van~Eerde, Rutte, and
  Roe}]{claessens2007review}
Brigitte~JC Claessens, Wendelien Van~Eerde, Christel~G Rutte, and Robert~A Roe.
  2007.
\newblock A review of the time management literature.
\newblock \emph{Personnel review}.

\bibitem[{Clark et~al.(2021)Clark, Salvador, Schwenk, Bonafilia, Yatskar,
  Kolve, Herrasti, Choi, Mehta, Skjonsberg et~al.}]{clark2021iconary}
Christopher Clark, Jordi Salvador, Dustin Schwenk, Derrick Bonafilia, Mark
  Yatskar, Eric Kolve, Alvaro Herrasti, Jonghyun Choi, Sachin Mehta, Sam
  Skjonsberg, et~al. 2021.
\newblock Iconary: A pictionary-based game for testing multimodal communication
  with drawings and text.
\newblock In \emph{Proceedings of the 2021 Conference on Empirical Methods in
  Natural Language Processing}, pages 1864--1886.

\bibitem[{Devlin et~al.(2019)Devlin, Chang, Lee, and
  Toutanova}]{devlin-etal-2019-bert}
Jacob Devlin, Ming-Wei Chang, Kenton Lee, and Kristina Toutanova. 2019.
\newblock \href {https://doi.org/10.18653/v1/N19-1423} {{BERT}: Pre-training of
  deep bidirectional transformers for language understanding}.
\newblock In \emph{Proceedings of the 2019 Conference of the North {A}merican
  Chapter of the Association for Computational Linguistics: Human Language
  Technologies, Volume 1 (Long and Short Papers)}, pages 4171--4186,
  Minneapolis, Minnesota. Association for Computational Linguistics.

\bibitem[{Dudy et~al.(2021)Dudy, Bedrick, and
  Webber}]{dudy-etal-2021-refocusing}
Shiran Dudy, Steven Bedrick, and Bonnie Webber. 2021.
\newblock \href {https://doi.org/10.18653/v1/2021.emnlp-main.421} {Refocusing
  on relevance: Personalization in {NLG}}.
\newblock In \emph{Proceedings of the 2021 Conference on Empirical Methods in
  Natural Language Processing}, pages 5190--5202, Online and Punta Cana,
  Dominican Republic. Association for Computational Linguistics.

\bibitem[{Efrat et~al.(2021)Efrat, Shaham, Kilman, and
  Levy}]{efrat2021cryptonite}
Avia Efrat, Uri Shaham, Dan Kilman, and Omer Levy. 2021.
\newblock Cryptonite: A cryptic crossword benchmark for extreme ambiguity in
  language.
\newblock In \emph{Proceedings of the 2021 Conference on Empirical Methods in
  Natural Language Processing}, pages 4186--4192.

\bibitem[{Falke and Lehnen(2021)}]{falke2021feedback}
Tobias Falke and Patrick Lehnen. 2021.
\newblock Feedback attribution for counterfactual bandit learning in
  multi-domain spoken language understanding.
\newblock In \emph{Proceedings of the 2021 Conference on Empirical Methods in
  Natural Language Processing}, pages 1190--1198.

\bibitem[{Flek(2020)}]{flek2020returning}
Lucie Flek. 2020.
\newblock Returning the n to nlp: Towards contextually personalized
  classification models.
\newblock In \emph{Proceedings of the 58th annual meeting of the association
  for computational linguistics}, pages 7828--7838.

\bibitem[{Garg and Moschitti(2021)}]{garg2021will}
Siddhant Garg and Alessandro Moschitti. 2021.
\newblock Will this question be answered? question filtering via answer model
  distillation for efficient question answering.
\newblock In \emph{Proceedings of the 2021 Conference on Empirical Methods in
  Natural Language Processing}, pages 7329--7346.

\bibitem[{Gehrmann et~al.(2021)Gehrmann, Adewumi, Aggarwal, Ammanamanchi,
  Aremu, Bosselut, Chandu, Clinciu, Das, Dhole, Du, Durmus, Du{\v{s}}ek,
  Emezue, Gangal, Garbacea, Hashimoto, Hou, Jernite, Jhamtani, Ji, Jolly, Kale,
  Kumar, Ladhak, Madaan, Maddela, Mahajan, Mahamood, Majumder, Martins,
  McMillan-Major, Mille, van Miltenburg, Nadeem, Narayan, Nikolaev,
  Niyongabo~Rubungo, Osei, Parikh, Perez-Beltrachini, Rao, Raunak, Rodriguez,
  Santhanam, Sedoc, Sellam, Shaikh, Shimorina, Sobrevilla~Cabezudo, Strobelt,
  Subramani, Xu, Yang, Yerukola, and Zhou}]{gehrmann-etal-2021-gem}
Sebastian Gehrmann, Tosin Adewumi, Karmanya Aggarwal, Pawan~Sasanka
  Ammanamanchi, Anuoluwapo Aremu, Antoine Bosselut, Khyathi~Raghavi Chandu,
  Miruna-Adriana Clinciu, Dipanjan Das, Kaustubh Dhole, Wanyu Du, Esin Durmus,
  Ond{\v{r}}ej Du{\v{s}}ek, Chris~Chinenye Emezue, Varun Gangal, Cristina
  Garbacea, Tatsunori Hashimoto, Yufang Hou, Yacine Jernite, Harsh Jhamtani,
  Yangfeng Ji, Shailza Jolly, Mihir Kale, Dhruv Kumar, Faisal Ladhak, Aman
  Madaan, Mounica Maddela, Khyati Mahajan, Saad Mahamood, Bodhisattwa~Prasad
  Majumder, Pedro~Henrique Martins, Angelina McMillan-Major, Simon Mille, Emiel
  van Miltenburg, Moin Nadeem, Shashi Narayan, Vitaly Nikolaev, Andre
  Niyongabo~Rubungo, Salomey Osei, Ankur Parikh, Laura Perez-Beltrachini,
  Niranjan~Ramesh Rao, Vikas Raunak, Juan~Diego Rodriguez, Sashank Santhanam,
  Jo{\~a}o Sedoc, Thibault Sellam, Samira Shaikh, Anastasia Shimorina,
  Marco~Antonio Sobrevilla~Cabezudo, Hendrik Strobelt, Nishant Subramani, Wei
  Xu, Diyi Yang, Akhila Yerukola, and Jiawei Zhou. 2021.
\newblock \href {https://doi.org/10.18653/v1/2021.gem-1.10} {The {GEM}
  benchmark: Natural language generation, its evaluation and metrics}.
\newblock In \emph{Proceedings of the 1st Workshop on Natural Language
  Generation, Evaluation, and Metrics (GEM 2021)}, pages 96--120, Online.
  Association for Computational Linguistics.

\bibitem[{Gehrmann et~al.(2022)Gehrmann, Bhattacharjee, Mahendiran, Wang,
  Papangelis, Madaan, McMillan-Major, Shvets, Upadhyay, Yao, Wilie,
  Bhagavatula, You, Thomson, Garbacea, Wang, Deutsch, Xiong, Jin, Gkatzia,
  Radev, Clark, Durmus, Ladhak, Ginter, Winata, Strobelt, Hayashi, Novikova,
  Kanerva, Chim, Zhou, Clive, Maynez, Sedoc, Juraska, Dhole, Chandu, Ribeiro,
  Tunstall, Zhang, Pushkarna, Creutz, White, Kale, Eddine, Daheim, Subramani,
  Dusek, Liang, Ammanamanchi, Zhu, Puduppully, Kriz, Shahriyar, Cardenas,
  Mahamood, Osei, Cahyawijaya, Štajner, Montella, {Shailza}, Jolly, Mille,
  Hasan, Shen, Adewumi, Raunak, Raheja, Nikolaev, Tsai, Jernite, Xu, Sang, Liu,
  and Hou}]{gemv2}
Sebastian Gehrmann, Abhik Bhattacharjee, Abinaya Mahendiran, Alex Wang,
  Alexandros Papangelis, Aman Madaan, Angelina McMillan-Major, Anna Shvets,
  Ashish Upadhyay, Bingsheng Yao, Bryan Wilie, Chandra Bhagavatula, Chaobin
  You, Craig Thomson, Cristina Garbacea, Dakuo Wang, Daniel Deutsch, Deyi
  Xiong, Di~Jin, Dimitra Gkatzia, Dragomir Radev, Elizabeth Clark, Esin Durmus,
  Faisal Ladhak, Filip Ginter, Genta~Indra Winata, Hendrik Strobelt, Hiroaki
  Hayashi, Jekaterina Novikova, Jenna Kanerva, Jenny Chim, Jiawei Zhou, Jordan
  Clive, Joshua Maynez, João Sedoc, Juraj Juraska, Kaustubh Dhole,
  Khyathi~Raghavi Chandu, Leonardo F.~R. Ribeiro, Lewis Tunstall, Li~Zhang,
  Mahima Pushkarna, Mathias Creutz, Michael White, Mihir~Sanjay Kale,
  Moussa~Kamal Eddine, Nico Daheim, Nishant Subramani, Ondrej Dusek, Paul~Pu
  Liang, Pawan~Sasanka Ammanamanchi, Qi~Zhu, Ratish Puduppully, Reno Kriz,
  Rifat Shahriyar, Ronald Cardenas, Saad Mahamood, Salomey Osei, Samuel
  Cahyawijaya, Sanja Štajner, Sebastien Montella, {Shailza}, Shailza Jolly,
  Simon Mille, Tahmid Hasan, Tianhao Shen, Tosin Adewumi, Vikas Raunak, Vipul
  Raheja, Vitaly Nikolaev, Vivian Tsai, Yacine Jernite, Ying Xu, Yisi Sang,
  Yixin Liu, and Yufang Hou. 2022.
\newblock {GEM}v2: {M}ultilingual {NLG} benchmarking in a single line of code.
\newblock \emph{arXiv preprint arXiv:2206.11249}.

\bibitem[{Gerz et~al.(2021)Gerz, Su, Kusztos, Mondal, Lis, Singhal,
  Mrk{\v{s}}i{\'c}, Wen, and Vuli{\'c}}]{gerz2021multilingual}
Daniela Gerz, Pei-Hao Su, Razvan Kusztos, Avishek Mondal, Micha{\l} Lis, Eshan
  Singhal, Nikola Mrk{\v{s}}i{\'c}, Tsung-Hsien Wen, and Ivan Vuli{\'c}. 2021.
\newblock Multilingual and cross-lingual intent detection from spoken data.
\newblock In \emph{Proceedings of the 2021 Conference on Empirical Methods in
  Natural Language Processing}, pages 7468--7475.

\bibitem[{Greenwald and Banaji(1995)}]{greenwald1995implicit}
Anthony~G Greenwald and Mahzarin~R Banaji. 1995.
\newblock Implicit social cognition: attitudes, self-esteem, and stereotypes.
\newblock \emph{Psychological review}, 102(1):4.

\bibitem[{Gu et~al.(2021)Gu, Ling, Wu, Liu, Chen, and Zhu}]{gu2021detecting}
Jia-Chen Gu, Zhenhua Ling, Yu~Wu, Quan Liu, Zhigang Chen, and Xiaodan Zhu.
  2021.
\newblock Detecting speaker personas from conversational texts.
\newblock In \emph{Proceedings of the 2021 Conference on Empirical Methods in
  Natural Language Processing}, pages 1126--1136.

\bibitem[{Hochreiter and Schmidhuber(1997)}]{6795963}
Sepp Hochreiter and Jürgen Schmidhuber. 1997.
\newblock \href {https://doi.org/10.1162/neco.1997.9.8.1735} {Long short-term
  memory}.
\newblock \emph{Neural Computation}, 9(8):1735--1780.

\bibitem[{Huang et~al.(2021)Huang, He, and Liu}]{huang2021improving}
Canming Huang, Weinan He, and Yongmei Liu. 2021.
\newblock Improving unsupervised commonsense reasoning using knowledge-enabled
  natural language inference.
\newblock In \emph{Findings of the Association for Computational Linguistics:
  EMNLP 2021}, pages 4875--4885.

\bibitem[{Jhamtani et~al.(2021)Jhamtani, Gangal, Hovy, and
  Berg-Kirkpatrick}]{jhamtani2021investigating}
Harsh Jhamtani, Varun Gangal, Eduard Hovy, and Taylor Berg-Kirkpatrick. 2021.
\newblock Investigating robustness of dialog models to popular figurative
  language constructs.
\newblock In \emph{Proceedings of the 2021 Conference on Empirical Methods in
  Natural Language Processing}, pages 7476--7485.

\bibitem[{Johnson(2021)}]{Khari2021the}
Khari Johnson. 2021.
\newblock The efforts to make text-based ai less racist and terrible.
\newblock \url{https://tinyurl.com/5x8rah4s}.
\newblock Accessed: 17 June 2021.

\bibitem[{Kahardipraja et~al.(2021)Kahardipraja, Madureira, and
  Schlangen}]{kahardipraja2021towards}
Patrick Kahardipraja, Brielen Madureira, and David Schlangen. 2021.
\newblock Towards incremental transformers: An empirical analysis of
  transformer models for incremental nlu.
\newblock In \emph{Proceedings of the 2021 Conference on Empirical Methods in
  Natural Language Processing}, pages 1178--1189.

\bibitem[{Kalyan et~al.(2021)Kalyan, Kumar, Chandrasekaran, Sabharwal, and
  Clark}]{kalyan2021much}
Ashwin Kalyan, Abhinav Kumar, Arjun Chandrasekaran, Ashish Sabharwal, and Peter
  Clark. 2021.
\newblock How much coffee was consumed during emnlp 2019? fermi problems: A new
  reasoning challenge for ai.
\newblock In \emph{Proceedings of the 2021 Conference on Empirical Methods in
  Natural Language Processing}, pages 7318--7328.

\bibitem[{Kann and
  Monsalve-Mercado(2021)}]{kann-monsalve-mercado-2021-coloring}
Katharina Kann and Mauro~M. Monsalve-Mercado. 2021.
\newblock \href {https://doi.org/10.18653/v1/2021.eacl-main.230} {Coloring the
  black box: What synesthesia tells us about character embeddings}.
\newblock In \emph{Proceedings of the 16th Conference of the European Chapter
  of the Association for Computational Linguistics: Main Volume}, pages
  2673--2685, Online. Association for Computational Linguistics.

\bibitem[{Lakei(1973)}]{lakei1973get}
A~Lakei. 1973.
\newblock How to get control of your time and life.
\newblock \emph{New York: Nal Penguin Inc}.

\bibitem[{Lavi et~al.(2021)Lavi, Rabinovich, Shlomov, Boaz, Ronen, and
  Tavor}]{lavi2021we}
Ofer Lavi, Ella Rabinovich, Segev Shlomov, David Boaz, Inbal Ronen, and
  Ateret~Anaby Tavor. 2021.
\newblock We’ve had this conversation before: A novel approach to measuring
  dialog similarity.
\newblock In \emph{Proceedings of the 2021 Conference on Empirical Methods in
  Natural Language Processing}, pages 1169--1177.

\bibitem[{Liang et~al.(2021)Liang, Zhou, Meng, Xu, Chen, Su, and
  Zhou}]{liang2021towards}
Yunlong Liang, Chulun Zhou, Fandong Meng, Jinan Xu, Yufeng Chen, Jinsong Su,
  and Jie Zhou. 2021.
\newblock Towards making the most of dialogue characteristics for neural chat
  translation.
\newblock In \emph{Proceedings of the 2021 Conference on Empirical Methods in
  Natural Language Processing}, pages 67--79.

\bibitem[{Lipton and Steinhardt(2019)}]{lipton-steinhart}
Zachary~C. Lipton and Jacob Steinhardt. 2019.
\newblock \href {https://doi.org/10.1145/3317287.3328534} {Troubling trends in
  machine learning scholarship: Some ml papers suffer from flaws that could
  mislead the public and stymie future research.}
\newblock \emph{Queue}, 17(1):45–77.

\bibitem[{Liu et~al.(2021)Liu, Du, Li, Li, and Chen}]{liu2021cross}
Dan Liu, Mengge Du, Xiaoxi Li, Ya~Li, and Enhong Chen. 2021.
\newblock Cross attention augmented transducer networks for simultaneous
  translation.
\newblock In \emph{Proceedings of the 2021 Conference on Empirical Methods in
  Natural Language Processing}, pages 39--55.

\bibitem[{Ma et~al.(2021)Ma, Takanobu, and Huang}]{ma2021cr}
Wenchang Ma, Ryuichi Takanobu, and Minlie Huang. 2021.
\newblock Cr-walker: Tree-structured graph reasoning and dialog acts for
  conversational recommendation.
\newblock In \emph{Proceedings of the 2021 Conference on Empirical Methods in
  Natural Language Processing}, pages 1839--1851.

\bibitem[{Madotto et~al.(2021)Madotto, Lin, Zhou, Moon, Crook, Liu, Yu, Cho,
  Fung, and Wang}]{madotto2021continual}
Andrea Madotto, Zhaojiang Lin, Zhenpeng Zhou, Seungwhan Moon, Paul~A Crook,
  Bing Liu, Zhou Yu, Eunjoon Cho, Pascale Fung, and Zhiguang Wang. 2021.
\newblock Continual learning in task-oriented dialogue systems.
\newblock In \emph{Proceedings of the 2021 Conference on Empirical Methods in
  Natural Language Processing}, pages 7452--7467.

\bibitem[{Major et~al.(2002)Major, Klein, and Ehrhart}]{major2002work}
Virginia~Smith Major, Katherine~J Klein, and Mark~G Ehrhart. 2002.
\newblock Work time, work interference with family, and psychological distress.
\newblock \emph{Journal of applied psychology}, 87(3):427.

\bibitem[{Moghe et~al.(2021)Moghe, Steedman, and Birch}]{moghe2021cross}
Nikita Moghe, Mark Steedman, and Alexandra Birch. 2021.
\newblock Cross-lingual intermediate fine-tuning improves dialogue state
  tracking.
\newblock In \emph{Proceedings of the 2021 Conference on Empirical Methods in
  Natural Language Processing}, pages 1137--1150.

\bibitem[{Moster et~al.(2021)Moster, Ford, and Rodeghero}]{moster2021my}
Makayla Moster, Denae Ford, and Paige Rodeghero. 2021.
\newblock " is my mic on?" preparing se students for collaborative remote work
  and hybrid team communication.
\newblock In \emph{2021 IEEE/ACM 43rd International Conference on Software
  Engineering: Software Engineering Education and Training (ICSE-SEET)}, pages
  89--94. IEEE.

\bibitem[{Nan et~al.(2021)Nan, Radev, Zhang, Rau, Sivaprasad, Hsieh, Tang,
  Vyas, Verma, Krishna, Liu, Irwanto, Pan, Rahman, Zaidi, Mutuma, Tarabar,
  Gupta, Yu, Tan, Lin, Xiong, Socher, and Rajani}]{nan-etal-2021-dart}
Linyong Nan, Dragomir Radev, Rui Zhang, Amrit Rau, Abhinand Sivaprasad,
  Chiachun Hsieh, Xiangru Tang, Aadit Vyas, Neha Verma, Pranav Krishna,
  Yangxiaokang Liu, Nadia Irwanto, Jessica Pan, Faiaz Rahman, Ahmad Zaidi,
  Mutethia Mutuma, Yasin Tarabar, Ankit Gupta, Tao Yu, Yi~Chern Tan,
  Xi~Victoria Lin, Caiming Xiong, Richard Socher, and Nazneen~Fatema Rajani.
  2021.
\newblock \href {https://doi.org/10.18653/v1/2021.naacl-main.37} {{DART}:
  Open-domain structured data record to text generation}.
\newblock In \emph{Proceedings of the 2021 Conference of the North American
  Chapter of the Association for Computational Linguistics: Human Language
  Technologies}, pages 432--447, Online. Association for Computational
  Linguistics.

\bibitem[{Ni et~al.(2021)Ni, Smith, Yuan, Larivière, and
  Sugimoto}]{doi:10.1126/sciadv.abe4639}
Chaoqun Ni, Elise Smith, Haimiao Yuan, Vincent Larivière, and Cassidy~R.
  Sugimoto. 2021.
\newblock \href {https://doi.org/10.1126/sciadv.abe4639} {The gendered nature
  of authorship}.
\newblock \emph{Science Advances}, 7(36):eabe4639.

\bibitem[{Ouyang et~al.(2021)Ouyang, Wang, Pang, Sun, Tian, Wu, and
  Wang}]{ouyang2021ernie}
Xuan Ouyang, Shuohuan Wang, Chao Pang, Yu~Sun, Hao Tian, Hua Wu, and Haifeng
  Wang. 2021.
\newblock Ernie-m: Enhanced multilingual representation by aligning
  cross-lingual semantics with monolingual corpora.
\newblock In \emph{Proceedings of the 2021 Conference on Empirical Methods in
  Natural Language Processing}, pages 27--38.

\bibitem[{Parcalabescu et~al.(2021)Parcalabescu, Trost, and
  Frank}]{parcalabescu-etal-2021-multimodality}
Letitia Parcalabescu, Nils Trost, and Anette Frank. 2021.
\newblock \href {https://aclanthology.org/2021.mmsr-1.1} {What is
  multimodality?}
\newblock In \emph{Proceedings of the 1st Workshop on Multimodal Semantic
  Representations (MMSR)}, pages 1--10, Groningen, Netherlands (Online).
  Association for Computational Linguistics.

\bibitem[{Raghu et~al.(2021)Raghu, Agarwal, Joshi et~al.}]{raghu2021end}
Dinesh Raghu, Shantanu Agarwal, Sachindra Joshi, et~al. 2021.
\newblock End-to-end learning of flowchart grounded task-oriented dialogs.
\newblock In \emph{Proceedings of the 2021 Conference on Empirical Methods in
  Natural Language Processing}, pages 4348--4366.

\bibitem[{Raji et~al.(2022)Raji, Kumar, Horowitz, and Selbst}]{raji2022fallacy}
Inioluwa~Deborah Raji, I~Elizabeth Kumar, Aaron Horowitz, and Andrew Selbst.
  2022.
\newblock The fallacy of ai functionality.
\newblock In \emph{2022 ACM Conference on Fairness, Accountability, and
  Transparency}, pages 959--972.

\bibitem[{Rodriguez et~al.(2021)Rodriguez, Barrow, Hoyle, Lalor, Jia, and
  Boyd-Graber}]{rodriguez-etal-2021-evaluation}
Pedro Rodriguez, Joe Barrow, Alexander~Miserlis Hoyle, John~P. Lalor, Robin
  Jia, and Jordan Boyd-Graber. 2021.
\newblock \href {https://doi.org/10.18653/v1/2021.acl-long.346} {Evaluation
  examples are not equally informative: How should that change {NLP}
  leaderboards?}
\newblock In \emph{Proceedings of the 59th Annual Meeting of the Association
  for Computational Linguistics and the 11th International Joint Conference on
  Natural Language Processing (Volume 1: Long Papers)}, pages 4486--4503,
  Online. Association for Computational Linguistics.

\bibitem[{Rogers(2021)}]{rogers-2021-changing}
Anna Rogers. 2021.
\newblock \href {https://doi.org/10.18653/v1/2021.acl-long.170} {Changing the
  world by changing the data}.
\newblock In \emph{Proceedings of the 59th Annual Meeting of the Association
  for Computational Linguistics and the 11th International Joint Conference on
  Natural Language Processing (Volume 1: Long Papers)}, pages 2182--2194,
  Online. Association for Computational Linguistics.

\bibitem[{Salesky et~al.(2021)Salesky, Etter, and Post}]{salesky2021robust}
Elizabeth Salesky, David Etter, and Matt Post. 2021.
\newblock Robust open-vocabulary translation from visual text representations.
\newblock In \emph{Proceedings of the 2021 Conference on Empirical Methods in
  Natural Language Processing}, pages 7235--7252.

\bibitem[{Simon(2020)}]{Simon2020google}
Simon. 2020.
\newblock Google duplex: The effects of deception on well-being.
\newblock \url{https://tinyurl.com/2yadfuer}.
\newblock Accessed: 11 June 2020.

\bibitem[{Song et~al.(2021)Song, Kim, and Yoon}]{song2021alignart}
Jongyoon Song, Sungwon Kim, and Sungroh Yoon. 2021.
\newblock Alignart: Non-autoregressive neural machine translation by jointly
  learning to estimate alignment and translate.
\newblock In \emph{Proceedings of the 2021 Conference on Empirical Methods in
  Natural Language Processing}, pages 1--14.

\bibitem[{Srivastava et~al.(2022)Srivastava, Rastogi, Rao, Shoeb, Abid, Fisch,
  Brown, Santoro, Gupta, Garriga-Alonso, Kluska, Lewkowycz, Agarwal, Power,
  Ray, Warstadt, Kocurek, Safaya, Tazarv, Xiang, Parrish, Nie, Hussain, Askell,
  Dsouza, Slone, Rahane, Iyer, Andreassen, Madotto, Santilli, Stuhlmüller,
  Dai, La, Lampinen, Zou, Jiang, Chen, Vuong, Gupta, Gottardi, Norelli,
  Venkatesh, Gholamidavoodi, Tabassum, Menezes, Kirubarajan, Mullokandov,
  Sabharwal, Herrick, Efrat, Erdem, Karakaş, Roberts, Loe, Zoph, Bojanowski,
  Özyurt, Hedayatnia, Neyshabur, Inden, Stein, Ekmekci, Lin, Howald, Diao,
  Dour, Stinson, Argueta, Ramírez, Singh, Rathkopf, Meng, Baral, Wu,
  Callison-Burch, Waites, Voigt, Manning, Potts, Ramirez, Rivera, Siro, Raffel,
  Ashcraft, Garbacea, Sileo, Garrette, Hendrycks, Kilman, Roth, Freeman,
  Khashabi, Levy, González, Perszyk, Hernandez, Chen, Ippolito, Gilboa, Dohan,
  Drakard, Jurgens, Datta, Ganguli, Emelin, Kleyko, Yuret, Chen, Tam, Hupkes,
  Misra, Buzan, Mollo, Yang, Lee, Shutova, Cubuk, Segal, Hagerman, Barnes,
  Donoway, Pavlick, Rodola, Lam, Chu, Tang, Erdem, Chang, Chi, Dyer, Jerzak,
  Kim, Manyasi, Zheltonozhskii, Xia, Siar, Martínez-Plumed, Happé, Chollet,
  Rong, Mishra, Winata, de~Melo, Kruszewski, Parascandolo, Mariani, Wang,
  Jaimovitch-López, Betz, Gur-Ari, Galijasevic, Kim, Rashkin, Hajishirzi,
  Mehta, Bogar, Shevlin, Schütze, Yakura, Zhang, Wong, Ng, Noble, Jumelet,
  Geissinger, Kernion, Hilton, Lee, Fisac, Simon, Koppel, Zheng, Zou, Kocoń,
  Thompson, Kaplan, Radom, Sohl-Dickstein, Phang, Wei, Yosinski, Novikova,
  Bosscher, Marsh, Kim, Taal, Engel, Alabi, Xu, Song, Tang, Waweru, Burden,
  Miller, Balis, Berant, Frohberg, Rozen, Hernandez-Orallo, Boudeman, Jones,
  Tenenbaum, Rule, Chua, Kanclerz, Livescu, Krauth, Gopalakrishnan, Ignatyeva,
  Markert, Dhole, Gimpel, Omondi, Mathewson, Chiafullo, Shkaruta, Shridhar,
  McDonell, Richardson, Reynolds, Gao, Zhang, Dugan, Qin, Contreras-Ochando,
  Morency, Moschella, Lam, Noble, Schmidt, He, Colón, Metz, Şenel, Bosma,
  Sap, ter Hoeve, Farooqi, Faruqui, Mazeika, Baturan, Marelli, Maru, Quintana,
  Tolkiehn, Giulianelli, Lewis, Potthast, Leavitt, Hagen, Schubert,
  Baitemirova, Arnaud, McElrath, Yee, Cohen, Gu, Ivanitskiy, Starritt, Strube,
  Swędrowski, Bevilacqua, Yasunaga, Kale, Cain, Xu, Suzgun, Tiwari, Bansal,
  Aminnaseri, Geva, Gheini, T, Peng, Chi, Lee, Krakover, Cameron, Roberts,
  Doiron, Nangia, Deckers, Muennighoff, Keskar, Iyer, Constant, Fiedel, Wen,
  Zhang, Agha, Elbaghdadi, Levy, Evans, Casares, Doshi, Fung, Liang, Vicol,
  Alipoormolabashi, Liao, Liang, Chang, Eckersley, Htut, Hwang, Miłkowski,
  Patil, Pezeshkpour, Oli, Mei, Lyu, Chen, Banjade, Rudolph, Gabriel, Habacker,
  Delgado, Millière, Garg, Barnes, Saurous, Arakawa, Raymaekers, Frank,
  Sikand, Novak, Sitelew, LeBras, Liu, Jacobs, Zhang, Salakhutdinov, Chi, Lee,
  Stovall, Teehan, Yang, Singh, Mohammad, Anand, Dillavou, Shleifer, Wiseman,
  Gruetter, Bowman, Schoenholz, Han, Kwatra, Rous, Ghazarian, Ghosh, Casey,
  Bischoff, Gehrmann, Schuster, Sadeghi, Hamdan, Zhou, Srivastava, Shi, Singh,
  Asaadi, Gu, Pachchigar, Toshniwal, Upadhyay, Shyamolima, Shakeri, Thormeyer,
  Melzi, Reddy, Makini, Lee, Torene, Hatwar, Dehaene, Divic, Ermon, Biderman,
  Lin, Prasad, Piantadosi, Shieber, Misherghi, Kiritchenko, Mishra, Linzen,
  Schuster, Li, Yu, Ali, Hashimoto, Wu, Desbordes, Rothschild, Phan, Wang,
  Nkinyili, Schick, Kornev, Telleen-Lawton, Tunduny, Gerstenberg, Chang,
  Neeraj, Khot, Shultz, Shaham, Misra, Demberg, Nyamai, Raunak, Ramasesh,
  Prabhu, Padmakumar, Srikumar, Fedus, Saunders, Zhang, Vossen, Ren, Tong,
  Zhao, Wu, Shen, Yaghoobzadeh, Lakretz, Song, Bahri, Choi, Yang, Hao, Chen,
  Belinkov, Hou, Hou, Bai, Seid, Zhao, Wang, Wang, Wang, and
  Wu}]{srivastava2022beyond}
Aarohi Srivastava, Abhinav Rastogi, Abhishek Rao, Abu Awal~Md Shoeb, Abubakar
  Abid, Adam Fisch, Adam~R. Brown, Adam Santoro, Aditya Gupta, Adrià
  Garriga-Alonso, Agnieszka Kluska, Aitor Lewkowycz, Akshat Agarwal, Alethea
  Power, Alex Ray, Alex Warstadt, Alexander~W. Kocurek, Ali Safaya, Ali Tazarv,
  Alice Xiang, Alicia Parrish, Allen Nie, Aman Hussain, Amanda Askell, Amanda
  Dsouza, Ambrose Slone, Ameet Rahane, Anantharaman~S. Iyer, Anders Andreassen,
  Andrea Madotto, Andrea Santilli, Andreas Stuhlmüller, Andrew Dai, Andrew La,
  Andrew Lampinen, Andy Zou, Angela Jiang, Angelica Chen, Anh Vuong, Animesh
  Gupta, Anna Gottardi, Antonio Norelli, Anu Venkatesh, Arash Gholamidavoodi,
  Arfa Tabassum, Arul Menezes, Arun Kirubarajan, Asher Mullokandov, Ashish
  Sabharwal, Austin Herrick, Avia Efrat, Aykut Erdem, Ayla Karakaş, B.~Ryan
  Roberts, Bao~Sheng Loe, Barret Zoph, Bartłomiej Bojanowski, Batuhan Özyurt,
  Behnam Hedayatnia, Behnam Neyshabur, Benjamin Inden, Benno Stein, Berk
  Ekmekci, Bill~Yuchen Lin, Blake Howald, Cameron Diao, Cameron Dour, Catherine
  Stinson, Cedrick Argueta, César~Ferri Ramírez, Chandan Singh, Charles
  Rathkopf, Chenlin Meng, Chitta Baral, Chiyu Wu, Chris Callison-Burch, Chris
  Waites, Christian Voigt, Christopher~D. Manning, Christopher Potts, Cindy
  Ramirez, Clara~E. Rivera, Clemencia Siro, Colin Raffel, Courtney Ashcraft,
  Cristina Garbacea, Damien Sileo, Dan Garrette, Dan Hendrycks, Dan Kilman, Dan
  Roth, Daniel Freeman, Daniel Khashabi, Daniel Levy, Daniel~Moseguí
  González, Danielle Perszyk, Danny Hernandez, Danqi Chen, Daphne Ippolito,
  Dar Gilboa, David Dohan, David Drakard, David Jurgens, Debajyoti Datta, Deep
  Ganguli, Denis Emelin, Denis Kleyko, Deniz Yuret, Derek Chen, Derek Tam,
  Dieuwke Hupkes, Diganta Misra, Dilyar Buzan, Dimitri~Coelho Mollo, Diyi Yang,
  Dong-Ho Lee, Ekaterina Shutova, Ekin~Dogus Cubuk, Elad Segal, Eleanor
  Hagerman, Elizabeth Barnes, Elizabeth Donoway, Ellie Pavlick, Emanuele
  Rodola, Emma Lam, Eric Chu, Eric Tang, Erkut Erdem, Ernie Chang, Ethan~A.
  Chi, Ethan Dyer, Ethan Jerzak, Ethan Kim, Eunice~Engefu Manyasi, Evgenii
  Zheltonozhskii, Fanyue Xia, Fatemeh Siar, Fernando Martínez-Plumed,
  Francesca Happé, Francois Chollet, Frieda Rong, Gaurav Mishra, Genta~Indra
  Winata, Gerard de~Melo, Germán Kruszewski, Giambattista Parascandolo,
  Giorgio Mariani, Gloria Wang, Gonzalo Jaimovitch-López, Gregor Betz, Guy
  Gur-Ari, Hana Galijasevic, Hannah Kim, Hannah Rashkin, Hannaneh Hajishirzi,
  Harsh Mehta, Hayden Bogar, Henry Shevlin, Hinrich Schütze, Hiromu Yakura,
  Hongming Zhang, Hugh~Mee Wong, Ian Ng, Isaac Noble, Jaap Jumelet, Jack
  Geissinger, Jackson Kernion, Jacob Hilton, Jaehoon Lee, Jaime~Fernández
  Fisac, James~B. Simon, James Koppel, James Zheng, James Zou, Jan Kocoń, Jana
  Thompson, Jared Kaplan, Jarema Radom, Jascha Sohl-Dickstein, Jason Phang,
  Jason Wei, Jason Yosinski, Jekaterina Novikova, Jelle Bosscher, Jennifer
  Marsh, Jeremy Kim, Jeroen Taal, Jesse Engel, Jesujoba Alabi, Jiacheng Xu,
  Jiaming Song, Jillian Tang, Joan Waweru, John Burden, John Miller, John~U.
  Balis, Jonathan Berant, Jörg Frohberg, Jos Rozen, Jose Hernandez-Orallo,
  Joseph Boudeman, Joseph Jones, Joshua~B. Tenenbaum, Joshua~S. Rule, Joyce
  Chua, Kamil Kanclerz, Karen Livescu, Karl Krauth, Karthik Gopalakrishnan,
  Katerina Ignatyeva, Katja Markert, Kaustubh~D. Dhole, Kevin Gimpel, Kevin
  Omondi, Kory Mathewson, Kristen Chiafullo, Ksenia Shkaruta, Kumar Shridhar,
  Kyle McDonell, Kyle Richardson, Laria Reynolds, Leo Gao, Li~Zhang, Liam
  Dugan, Lianhui Qin, Lidia Contreras-Ochando, Louis-Philippe Morency, Luca
  Moschella, Lucas Lam, Lucy Noble, Ludwig Schmidt, Luheng He, Luis~Oliveros
  Colón, Luke Metz, Lütfi~Kerem Şenel, Maarten Bosma, Maarten Sap, Maartje
  ter Hoeve, Maheen Farooqi, Manaal Faruqui, Mantas Mazeika, Marco Baturan,
  Marco Marelli, Marco Maru, Maria Jose~Ramírez Quintana, Marie Tolkiehn,
  Mario Giulianelli, Martha Lewis, Martin Potthast, Matthew~L. Leavitt,
  Matthias Hagen, Mátyás Schubert, Medina~Orduna Baitemirova, Melody Arnaud,
  Melvin McElrath, Michael~A. Yee, Michael Cohen, Michael Gu, Michael
  Ivanitskiy, Michael Starritt, Michael Strube, Michał Swędrowski, Michele
  Bevilacqua, Michihiro Yasunaga, Mihir Kale, Mike Cain, Mimee Xu, Mirac
  Suzgun, Mo~Tiwari, Mohit Bansal, Moin Aminnaseri, Mor Geva, Mozhdeh Gheini,
  Mukund~Varma T, Nanyun Peng, Nathan Chi, Nayeon Lee, Neta Gur-Ari Krakover,
  Nicholas Cameron, Nicholas Roberts, Nick Doiron, Nikita Nangia, Niklas
  Deckers, Niklas Muennighoff, Nitish~Shirish Keskar, Niveditha~S. Iyer, Noah
  Constant, Noah Fiedel, Nuan Wen, Oliver Zhang, Omar Agha, Omar Elbaghdadi,
  Omer Levy, Owain Evans, Pablo Antonio~Moreno Casares, Parth Doshi, Pascale
  Fung, Paul~Pu Liang, Paul Vicol, Pegah Alipoormolabashi, Peiyuan Liao, Percy
  Liang, Peter Chang, Peter Eckersley, Phu~Mon Htut, Pinyu Hwang, Piotr
  Miłkowski, Piyush Patil, Pouya Pezeshkpour, Priti Oli, Qiaozhu Mei, Qing
  Lyu, Qinlang Chen, Rabin Banjade, Rachel~Etta Rudolph, Raefer Gabriel, Rahel
  Habacker, Ramón~Risco Del. 2022.
\newblock Beyond the imitation game: {Q}uantifying and extrapolating the
  capabilities of language models.
\newblock \emph{arXiv preprint arXiv:2206.04615}.

\bibitem[{Strubell et~al.(2019)Strubell, Ganesh, and
  McCallum}]{strubell2019energy}
Emma Strubell, Ananya Ganesh, and Andrew McCallum. 2019.
\newblock Energy and policy considerations for deep learning in nlp.
\newblock In \emph{Proceedings of the 57th Annual Meeting of the Association
  for Computational Linguistics}, pages 3645--3650.

\bibitem[{Van~Noorden(2018)}]{van2018leadership}
Richard Van~Noorden. 2018.
\newblock Leadership problems in the lab.
\newblock \emph{Nature}, 557(3).

\bibitem[{Vuli{\'c} et~al.(2021)Vuli{\'c}, Su, Coope, Gerz, Budzianowski,
  Casanueva, Mrk{\v{s}}i{\'c}, and Wen}]{vulic2021convfit}
Ivan Vuli{\'c}, Pei-Hao Su, Samuel Coope, Daniela Gerz, Pawe{\l} Budzianowski,
  I{\~n}igo Casanueva, Nikola Mrk{\v{s}}i{\'c}, and Tsung-Hsien Wen. 2021.
\newblock Convfit: Conversational fine-tuning of pretrained language models.
\newblock In \emph{Proceedings of the 2021 Conference on Empirical Methods in
  Natural Language Processing}, pages 1151--1168.

\bibitem[{Xu et~al.(2021)Xu, Van~Durme, and Murray}]{xu2021bert}
Haoran Xu, Benjamin Van~Durme, and Kenton Murray. 2021.
\newblock Bert, mbert, or bibert? a study on contextualized embeddings for
  neural machine translation.
\newblock In \emph{Proceedings of the 2021 Conference on Empirical Methods in
  Natural Language Processing}, pages 6663--6675.

\bibitem[{Zhang and Feng(2021)}]{zhang2021universal}
Shaolei Zhang and Yang Feng. 2021.
\newblock Universal simultaneous machine translation with mixture-of-experts
  wait-k policy.
\newblock In \emph{Proceedings of the 2021 Conference on Empirical Methods in
  Natural Language Processing}, pages 7306--7317.

\bibitem[{Zhang and Bansal(2021)}]{zhang2021finding}
Shiyue Zhang and Mohit Bansal. 2021.
\newblock Finding a balanced degree of automation for summary evaluation.
\newblock In \emph{Proceedings of the 2021 Conference on Empirical Methods in
  Natural Language Processing}, pages 6617--6632.

\bibitem[{Zhao et~al.(2021{\natexlab{a}})Zhao, Arthur, Haffari, Cohn, and
  Shareghi}]{zhao2021not}
Jinming Zhao, Philip Arthur, Gholamreza Haffari, Trevor Cohn, and Ehsan
  Shareghi. 2021{\natexlab{a}}.
\newblock It is not as good as you think! evaluating simultaneous machine
  translation on interpretation data.
\newblock In \emph{Proceedings of the 2021 Conference on Empirical Methods in
  Natural Language Processing}, pages 6707--6715.

\bibitem[{Zhao et~al.(2021{\natexlab{b}})Zhao, Wang, Zhu, and
  Wang}]{zhao2021efficient}
Yangyang Zhao, Zhenyu Wang, Changxi Zhu, and Shihan Wang. 2021{\natexlab{b}}.
\newblock Efficient dialogue complementary policy learning via deep q-network
  policy and episodic memory policy.
\newblock In \emph{Proceedings of the 2021 Conference on Empirical Methods in
  Natural Language Processing}, pages 4311--4323.

\bibitem[{Zhu et~al.(2021)Zhu, Gao, Guo, and Lou}]{zhu2021translating}
Kunrui Zhu, Yan Gao, Jiaqi Guo, and Jian-Guang Lou. 2021.
\newblock Translating headers of tabular data: A pilot study of schema
  translation.
\newblock In \emph{Proceedings of the 2021 Conference on Empirical Methods in
  Natural Language Processing}, pages 56--66.

\end{thebibliography}
\bibliographystyle{acl_natbib}

\clearpage
\appendix

\section{Appendix}
\label{sec:appendix}
In Table \ref{tab:emnlpanalysis} we provide the analysis conducted on selected EMNLP 2021 papers. \textit{Engaging with Users} indicates that researchers engage with humans, either during the design phase or for evaluation. In our analysis none of the papers engage with users throughout the process, leaving humans only to the evaluation part (3 papers). \textit{User-driven} indicates that the motivation is grounded in user needs (2 papers). The following tracks are considered: session 1: track A: Machine translation and multi-linguality 1, session 3: track B: Dialogue and interactive systems 1, session 4: track B: Dialogue and interactive systems 2, session 5: track A: question answering 1, session 6: track B: summarization, session 7: track A: machine translation and multi-linguality 2, session 7: track B: question answering 2. 

\begin{table*}
\centering
\small
\setlength{\tabcolsep}{5pt}
    \begin{tabular}{llll}
    \toprule
& \textbf{Paper} & \textbf{Engaging with Users} & \textbf{User-driven} \\ 
\midrule
1 & AligNART \citep{song2021alignart}  & no & no \\
2 & Zero-Shot Cross-Lingual Transfer \citep{chen2021zero}  & no & no \\
3 & ERNIE-M \citep{ouyang2021ernie} & no & no \\
4 & Cross attention augmented transducer \citep{liu2021cross} & no & no \\
5 & Translating Headers of Tabular Data \citep{zhu2021translating}  & no & no \\
6 & Towards Making the Most \citep{liang2021towards} & no & no \\
7 & MindCraft \citep{bara2021mindcraft} & yes & no \\
8 & Detecting Speaker Personas \citep{gu2021detecting} & no & no \\
9 & Cross-lingual Intermediate Fine-tuning \citep{moghe2021cross} & no & no \\
10 & ConvFiT \cite{vulic2021convfit} & no & no \\
11 & We’ve had this conversation before \citep{lavi2021we} & no & no \\
12 & Towards Incremental Transformers \citep{kahardipraja2021towards} & no & no \\
13 & Feedback Attribution \citep{falke2021feedback} & no & yes \\
14 & CR-Walker \citep{ma2021cr} & no & no \\
15 & Iconary \citep{clark2021iconary} & yes & no \\
16 & Improving Unsupervised Commonsense \citep{huang2021improving} & no & no \\
17 & Cryptonite \citep{efrat2021cryptonite} & no & no \\
18 & Efficient Dialogue Complementary Policy Learning \citep{zhao2021efficient} & yes & no \\
19 & End-to-End Learning of Flowchart \citep{raghu2021end} & no & yes \\
20 & Aspect-Controllable Opinion Summarization \citep{amplayo2021aspect} & no & no \\
21 & Finding a Balanced Degree of Automation \citep{zhang2021finding}  & no & no \\
22 & BERT, mBERT, or BiBERT \citep{xu2021bert} & no & no \\
23 & It Is Not As Good As You Think \citep{zhao2021not} & no & no \\
24 & Robust Open-Vocabulary Translation \citep{salesky2021robust} & no & no \\
25 & Universal Simultaneous Machine Translation \citep{zhang2021universal} & no & no \\
26 & How much coffee was consumed \citep{kalyan2021much} & no & no \\
27 & Will this Question be Answered \citep{garg2021will} & no & no \\
28 & Continual Learning \citep{madotto2021continual} & no & no \\
29 & Multilingual and Cross-Lingual Intent \citep{gerz2021multilingual} & no & no \\
30 & Investigating Robustness of Dialog Models \citep{jhamtani2021investigating} & no & no \\
    \bottomrule
    \end{tabular}
\caption{Our analysis of 30 randomly chosen papers from EMNLP 2021.}
\label{tab:emnlpanalysis}
\end{table*}

\end{document}